\documentclass[10pt, a4paper]{article}
\usepackage{lrec2022} 
\usepackage{multibib}
\newcites{languageresource}{Language Resources}
\usepackage{graphicx}
\usepackage{tabularx}
\usepackage{soul}
\usepackage{float}
\usepackage{booktabs}
\usepackage{hyperref}
\usepackage{balance}
\usepackage{titlesec}
\titleformat{\section}{\normalfont\large\bfseries\center}{\thesection.}{1em}{}
\titleformat{\subsection}{\normalfont\SmallTitleFont\bfseries\raggedright}{\thesubsection.}{1em}{}
\titleformat{\subsubsection}{\normalfont\normalsize\bfseries\raggedright}{\thesubsubsection.}{1em}{}
\renewcommand\thesection{\arabic{section}}
\renewcommand\thesubsection{\thesection.\arabic{subsection}}
\renewcommand\thesubsubsection{\thesubsection.\arabic{subsubsection}}

\usepackage{epstopdf}
\usepackage[utf8]{inputenc}
\usepackage{xstring}
\usepackage{color}

\title{A Multimodal German Dataset for Automatic Lip Reading Systems and Transfer Learning \\ \vspace*{.5\baselineskip} \normalfont{}}

\name{Gerald Schwiebert, Cornelius Weber, Leyuan Qu, Henrique Siqueira, Stefan Wermter} 

\address{Knowledge Technology, Department of Informatics,
University of Hamburg \\
         Vogt-Koelln-Str. 30, 22527 Hamburg \\
         g.schwiebert@mailbox.org\\
         \{weber, qu, siqueira, wermter\}@informatik.uni-hamburg.de\\}

\abstract{
Large datasets as required for deep learning of lip reading do not exist in many languages. In this paper we present the dataset GLips (German Lips) consisting of 250,000 publicly available videos of the faces of speakers of the Hessian Parliament, which was processed for word-level lip reading using an automatic pipeline. The format is similar to that of the English language LRW (Lip Reading in the Wild) dataset, with each video encoding one word of interest in a context of 1.16 seconds duration, which yields compatibility for studying transfer learning between both datasets. By training a deep neural network, we investigate whether lip reading has language-independent features, so that datasets of different languages can be used to improve lip reading models. We demonstrate learning from scratch and show that transfer learning from LRW to GLips and vice versa improves learning speed and performance, in particular for the validation set.
\\ \newline  }

\begin{document}
\maketitleabstract
\section{Introduction}

Lip reading is the ability of drawing conclusions about what is being said by visually observing a speaker's lips. In practice, however, it can rarely be considered on its own in human speech, since a variety of additional information is usually available in communication which, in combination with lip reading, can increase the information content of the incoming communication~\cite{klucharev2003electrophysiological}. These can consist of, for example, audio information, context, heuristics, gestures, facial expressions, or prior knowledge about what is being said. Also, special capabilities of the human brain and auditory sense allow us to apply filters that increase the focus on the desired communication, such as the well-known cocktail party effect~\cite{cherry1953some}. Thus, when phonemes, syllables, words, or entire sentences have an information deficit from the sender or receiver related to the communication pathway, lip reading can be one among many ways to compensate for this deficit. Ambient noise, hearing loss, slurred pronunciation, unfamiliar words, distance, or soundproof barriers are examples of communication problems between sender and receiver. Most people automatically look at the speaker's lips when intelligibility suffers, so they are all lip readers with varying degrees of skill~\cite{woodhouse2009review}.

There is a variety of possible applications for lip reading, such as disability support, sports communication in the press, voice control in noisy environments, additional accuracy for ASR systems, law enforcement and preservation of evidence. In recent years, enormous progress has been made in a number of technical areas, the effects of which enable an efficient technical evaluation of lip reading~\cite{paszke2019pytorch,chetlur2014cudnn}. High-quality cameras nowadays have high-resolution sensors whose light sensitivity, dynamic range and noise suppression in post-processing enable clear, sharp and detailed images. The memory capacity and computing power of graphic cards have increased to such an extent that complex computations are also possible on consumer PCs~\cite{lemley2017deep}. New deep learning models require large-scale labelled training data, which is not available for many languages yet. Our motivation is twofold:
\begin{enumerate}
    \item to contribute German data\nocitelanguageresource{GLips}\footnote{GLips is available at the following link: \url{https://www.inf.uni-hamburg.de/en/inst/ab/wtm/research/corpora.html}} for the generation of corpus-based lip reading models, and
    \item to evaluate this dataset by training a deep neural network and transfer learning to and from the English LRW dataset.
\end{enumerate}

\section{Background}

Datasets for lip reading are not yet as common as those for speech recognition.
In this section, 
we describe the details of the ``Lip Reading in the Wild" (LRW) dataset created by~\newcite{chung2016lip}, which is a popular benchmark and high-quality dataset for Automatic Lip Reading (ALR) and thematically related tasks such as Automatic Speech Recognition (ASR) in general. 
Other published datasets for lip reading are the Chinese LRW-1000 \cite{yang2019LRW1000} and the Romanian \cite{jitaru2020lrro}. We will use the English LRW for comparison and transfer learning.
In the second part of this section, we briefly address the particular situation of data protection and copyright in Germany and consider the practical obstacles that arise when creating large datasets.

\subsection{LRW – Lip Reading in the Wild}
The LRW dataset consists of short videos of people's faces uttering a defined word~\cite{chung2016lip}. Care was taken to ensure that the speakers' lips are clearly visible.

LRW is composed of MPEG4 videos, each 1.16s long and recorded at 25 frames per second (fps). The 500 classes of words, each with approximately 900-1100 instances, spoken by hundreds of different speakers were cut from BBC archives of news broadcasts, talk shows and interviews in 256$\times$256 pixels format and center-focused on the speaker via face detection. In each video clip, the word labelled as a directory name consisting of 5-12 letters is pronounced in a time-centred manner. Here, 50-word instances each are divided between validation and test directories, while the rest is reserved for training purposes. In addition, for each word there is a metadata file in .txt format containing BBC internal data such as disk reference, channel and program start. As an externally usable entry, the duration of the pronunciation of the respective word is noted in seconds. 
In our evaluation (cf.\ Section~\ref{chapter:evaluation}),
we cropped the videos of the LRW dataset to 96$\times$96 pixels focusing on the lips of the speakers to train our lip reading models.

\subsection{Copyright}
Copyright law (UrhG) in Germany and its related ancillary copyrights deal with the creation of works and the rights and powers of their creators. Videos that can be accessed and downloaded from publicly accessible platforms are in principle subject to §1 UrhG\footnote{§1 UrhG-Allgemeines-dejure.org: \url{https://dejure.org/gesetze/UrhG/1.html}} as a work as soon as they have a certain so-called creative level, i.e., they are enhanced, for example, by creative editing. Choosing such videos as a source for creating a dataset would involve a great deal of communication effort, as the permission of the author would have to be obtained in writing for each individual work. Videos from webcams or surveillance cameras without further significant creative editing generally lack this level of creativity, which is why they are particularly suitable as a data source for dataset creation, as long as the rules of the DSGVO\footnote{Datenschutz-Grundverordnung (DSGVO) - dejure.org: \url{https://dejure.org/gesetze/DSGVO}} are followed. Furthermore, in creating GLips, we comply with two special exceptions embedded into the German copyright law. First, we pursue a legitimate scientific interest for helping to enhance the support for the hearing impaired through the creation of our dataset and second, the politicians shown are persons of public interest recorded in a publically available parliament-recorded environment.

\subsection{General Data Protection Regulation in Germany (DSGVO)}
When people are recorded on film, these videos are subject to the DSGVO, which has been valid in the European Union since 25.05.2018. The DSGVO requirements cover, among other things, in Art.\ 5: 
(1) legality, (2) public interest for a specific purpose, (3) data minimization, (4) correctness, (5) storage time limit, and (6) integrity and confidentiality,
whereby special regulations regarding items 2 and 5 apply for scientific research purposes, which e.g., allow a longer storage period. This binding framework increases the bureaucratic, legal and possibly also personnel effort for a data protection officer, which further reduces the availability of datasets for machine learning purposes. But since biometric features are usually not mutable and we cannot yet fully estimate how many derivations from biometric data are possible in the future~\cite{faundez2005privacy}, we as dataset creators have a special responsibility with regard to copyright, data protection and licensing.

\section{GLips Dataset Creation}

\begin{figure}[h]
\centering
\includegraphics[width=0.9\columnwidth]{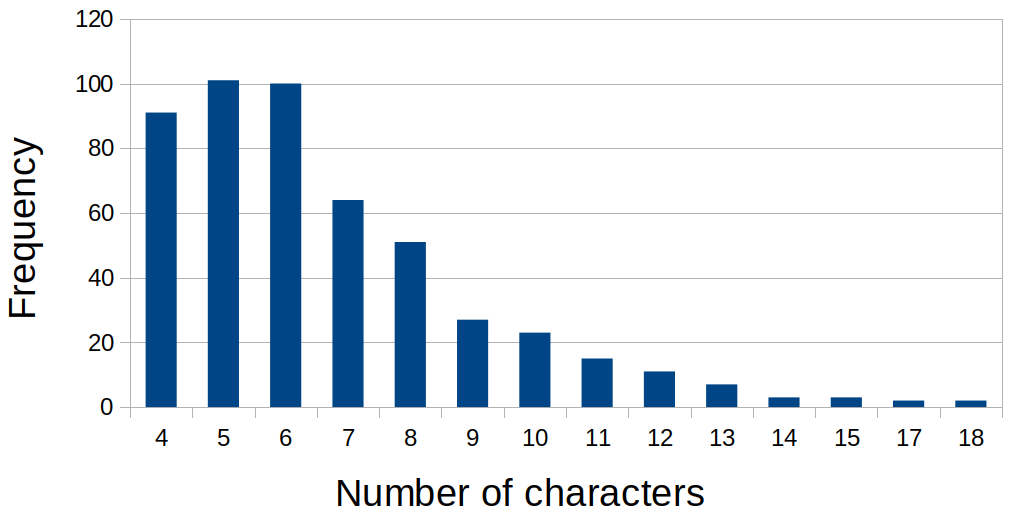} 
\caption{Word length distribution 
         in GLips}
\label{fig.1}
\end{figure}

\begin{figure*}[h]
\centering

\includegraphics[width=0.9\textwidth]{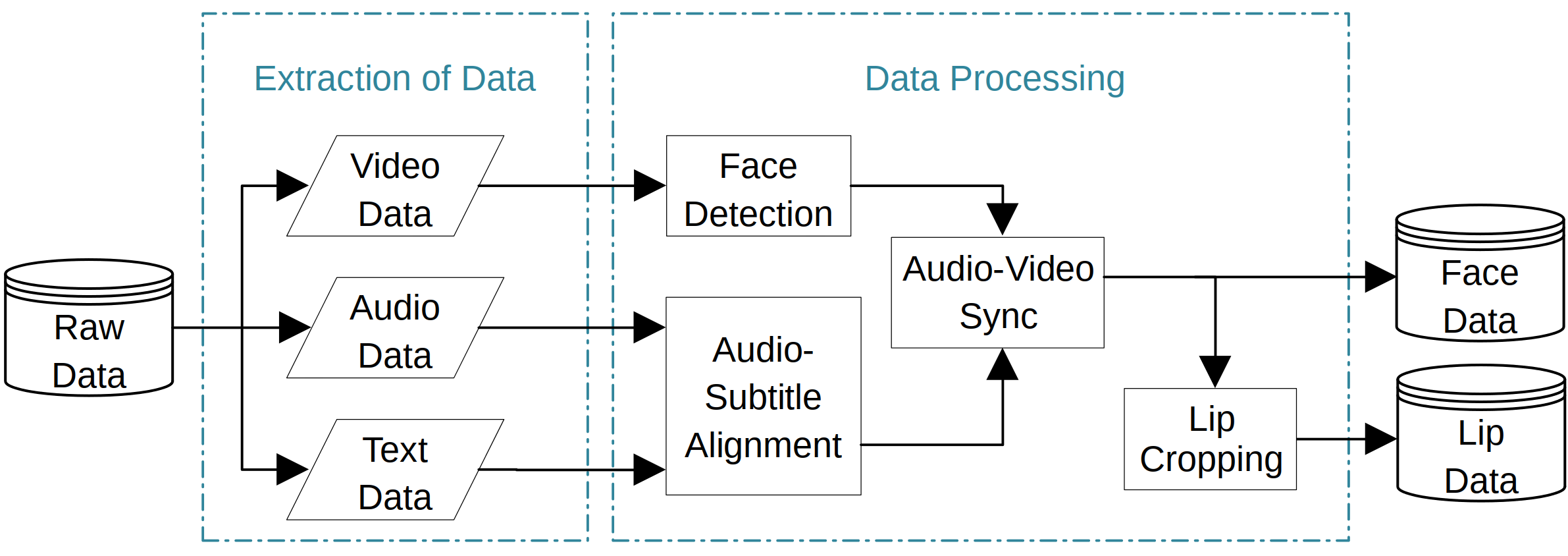} 
\caption{Pipeline for the generation of training data}
\label{fig.2}

\end{figure*}

This section gives an overview of the creation procedure of the dataset German Lips (GLips). Despite the focus of this paper and the name on the ALR domain, GLips should be applicable in the whole scientific ASR domain as versatile as possible, because, as explained in Section 2, the legally compliant creation of large video datasets in the German-speaking area is connected with some hurdles. Therefore, the creation of GLips is oriented towards LRW in order to ensure a high compatibility for methods such as transfer learning and experiments on the topic of Language Independence and to advance scientific knowledge in these areas. Furthermore, by choosing video material based on naturally spoken language in a natural environment, we decided to use this approach for ASR systems, as it produces more robust results for real-world applications than artificially generated datasets with as little noise as possible~\cite{burton2018speaker}.

GLips consists of 250,000 H264-compressed MPEG-4 videos of speakers' faces from parliamentary sessions of the Hessian Parliament, which are divided into 500 different words of 500 instances each. The word length distribution is shown in Fig.\ref{fig.1}. As with LRW, each video is 1.16s long at a frame rate of 25fps. The audio track was stored separately in an MPEG AAC audio file (.m4a). For each video there is an additional metadata textfile with the fields:
\begin{itemize}
\item Spoken word,
\item Start time of utterance in seconds,
\item End time of utterance in seconds,
\item Duration of utterance in seconds,
\item Corresponding numerical filename in the database.
\end{itemize}
Start- and end-time of utterance refers to the complete original video and not to the occurrence of the word in the clip.

\subsection{Acquisition}
With the permission of the Hessian Parliament, we used over 1000 videos and their respective subtitles. The Hessian parliament has published a superset of these videos also on its YouTube channel\footnote{YouTube - Hessischer Landtag: \url{https://www.youtube.com/c/HessischerLandtagOnline}}.
The subtitles are available as a separate text file and include manually created subtitles with time intervals. Similar to LRW subtitle editing~\cite{chung2016lip}, this leads to the issue that not all subtitles are verbatim, as in rare cases the content but not the exact spoken words have been reproduced in the subtitle, which means that despite checks, there are likely to be some words in the dataset that do not match the lip profile of the speaker. In order to create GLips, we also need the exact time of pronunciation and the duration of the utterance for each selected word. However, the subtitle files only contain one interval for each of several words. The solution to this problem via alignment using the WebMAUS service is discussed in section 3.3.

\subsection{Multimodal Processing Pipeline}

The technical creation of the multimodal dataset GLips is thematically divided into the two areas of extraction and processing of data. In Fig.~\ref{fig.2}, the entire pipeline is shown schematically from the existence of the raw data to the creation of training data suitable for machine learning, of which GLips represents a subset.

Since the original audio, video and subtitle data are already available in a separate form, the technical part of the data extraction is limited to the acquisition of all data and the cleaning of the text data from meta information so that only the spoken words are available as input for the next step.
The more complex part of the data processing is described in more detail in the following two sections and is divided into the two subsections audio subtitle alignment using WebMAUS and face detection. The audio and video files are synchronized in the last step, but are stored in separate files for the sake of more diverse processing options.

\vspace{5mm}
\begin{figure}[th]
\begin{center}
\includegraphics[width=\columnwidth]{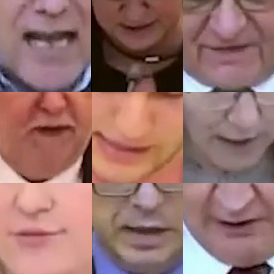} 

\caption{Example of GLips cropped to 96$\times$96 pixels}
\label{fig.3}
\end{center}
\end{figure}

The output of this pipeline consists of structured, processed, and augmented data suitable as potential training data for various areas of machine learning. The TextGrid files with their phonetic information, which are no longer needed by GLips, longer excerpts from aligned videos for sentence-based lip reading approaches, or video clips with several people to test attention mechanisms, are just a few ideas for research.
For the transfer learning in Section 4, we use a modified GLips dataset that was reduced in size from 256$\times$256 pixels to 96$\times$96 pixels (see Fig.~\ref{fig.3}) by additionally cropping the videos to focus on lip reading learning and to ensure better computability on consumer hardware.

\subsection{Subtitle Alignment using WebMAUS}

The Munich Automatic Segmentation System (MAUS) is a software for manifold speech data processing developed by~\newcite{epub13684} 
which is also available as a web service called WebMAUS\footnote{BAS web service interface: \url{https://clarin.phonetik.uni-muenchen.de/BASWebServices/interface}}~\cite{kisler2012signal,ide2017handbook}. For us, the modules G2P, Chunker and MAUS from this software package are of particular interest as an automatic pipeline via RESTful API to be able to create GLips. From our previously cleaned subtitle text files, a phonological transcript is created using G2P in BAS score format\footnote{Phonetik BAS: \url{https://www.phonetik.uni-muenchen.de/Bas/BasFormatsdeu.html}}, which is an open format for describing segmental information. A chunker is also needed to keep the audio and text information more easily computable, since some audio files come from hours of parliamentary sessions that can push the WebMAUS server to its capacity limits when sent in aggregate over several days. From these smaller segments, the aligner from the WebMAUS service can use the corresponding audio file to temporally match the text with the audio file in order to generate a TextGrid file with matching phonetic and word segments. There are three levels of analyzable information in this. As can be seen in Fig.~\ref{fig.4}, the levels ORT-MAU (orthographic information), KAN-MAU (canonical-phonemic word representation) and MAU (aligned transcription) including the corresponding time axis are available. With this information about an exact time period of the pronunciation of each word spoken in the audio file, it is possible to extract the temporally related video clip from the original video.

\subsection{Face Detection}

\begin{figure}[h]
\centering
\includegraphics[width=\columnwidth]{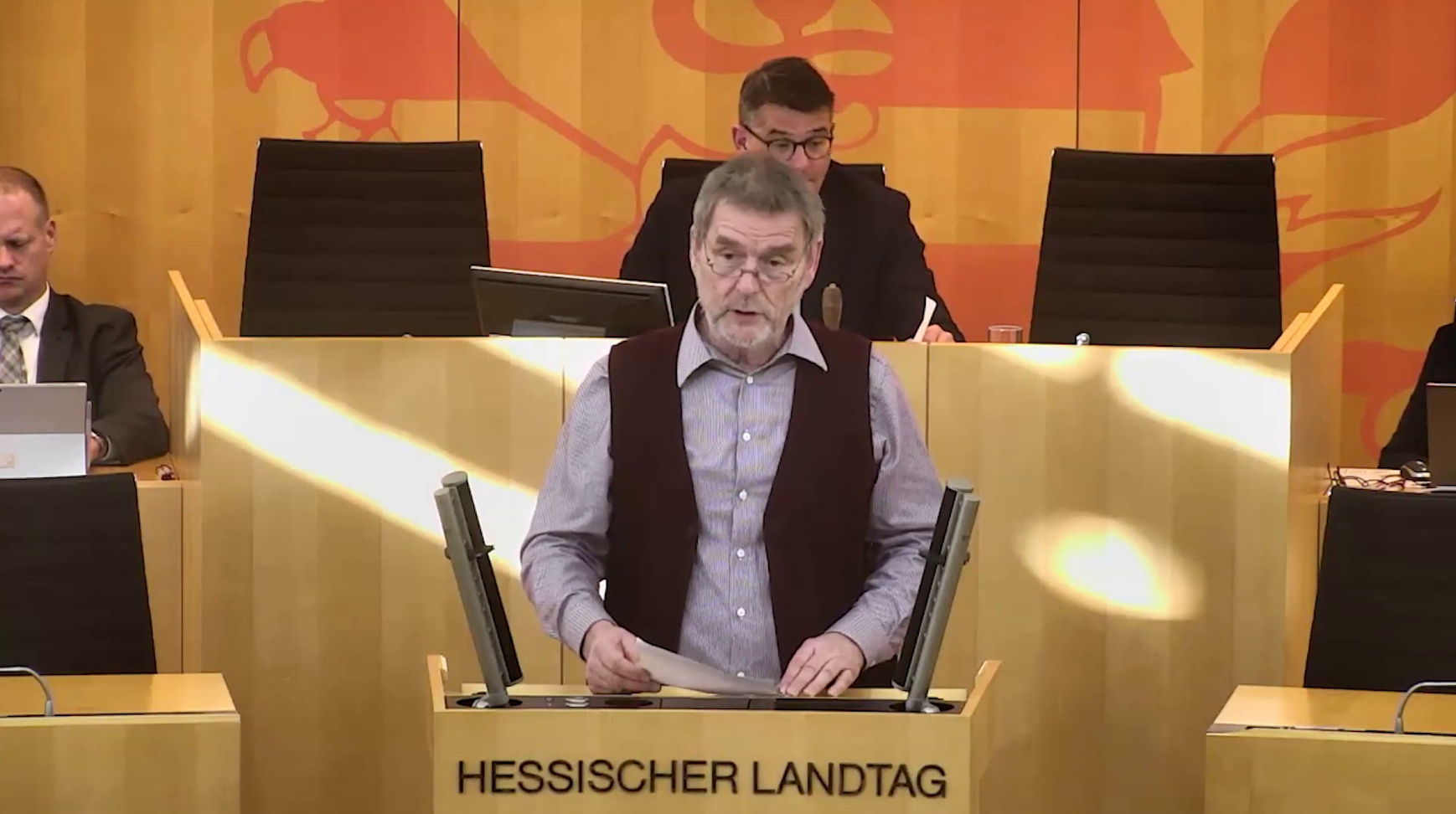} 

\caption{Example full screen view of the raw video data from the Hessian Parliament \vspace{5mm}}
\label{fig.5}
\end{figure}

The face detection was implemented based on the two Python libraries OpenCV\footnote{OpenCV: \url{https://opencv.org/}} with Nvidia-CUDA\footnote{CUDA Toolkit: \url{https://developer.nvidia.com/cuda-toolkit}} support for more efficient performance and face\_recognition\footnote{GitHub - ageitgey face\_recognition: \url{https://github.com/ageitgey/face_recognition}}. As shown in Fig.~\ref{fig.5}, the external conditions for speaker detection are almost perfectly suited to run face detection only on a section of the video due to the fixed podium and almost constant position of the webcam, which both reduces the processing time of all videos and makes attention mechanisms less important for speaker detection. Very few complications occur in the videos that these processing conditions do not satisfy, s.a. the speaker moving too vividly or being very tall which could cause the face detection to confuse the speakers face with the person sitting in the elevated position behind him.

\begin{figure*}[htbp]
\centering
\includegraphics[width=\textwidth]{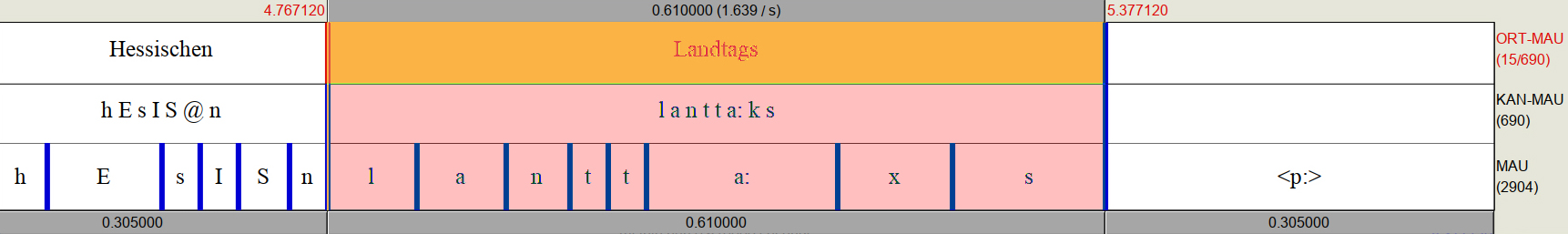} 
\caption{The spoken word ``Landtags" including context visualized from a TextGrid file using Praat software~\protect\cite{boersma2001praat}}
\label{fig.4}
\end{figure*}

\section{Comparison of GLips and LRW}

\begin{table}[tb]
\begin{tabular}{@{}p{26mm}p{18mm}p{18mm}@{}}
\toprule
\textbf{Video Properties}    & \textbf{LRW}    & \textbf{GLips}        \\ \midrule
Total number                 & $\sim$500,000      & 250,000                  \\
Number of different speakers & hundreds & $\sim$100             \\
Format                       & MPEG4           & MPEG4                 \\
Resolution in pixels                  & 256$\times$256        & 256$\times$256             \\
Length                       & 1,16s           & 1,16s                 \\
Framerate                    & 25fps           & 25fps                 \\
Word classes                 & 500             & 500                   \\
Instances                    & $\sim$1000      & 500                   \\
Word length in letters       & 5-12            & 4-18                  \\
Metadata file                & yes             & yes                   \\
Separate audio file          & no              & yes                   \\
TextGrid file                & no              & yes                   \\
Equipment level              & professional    & webcam                \\
Lighting                     & professional    & indoor standard       \\ \bottomrule
\end{tabular}
\caption{\label{tab:LRW vs. GLips}Comparison of LRW and GLips}
\end{table}

Both GLips and LRW contain large quantities of videos of speakers in frontal view perspectives to ensure a clear view on their lips. As seen in Table~\ref{tab:LRW vs. GLips}, special attention was paid to the aspect of compatibility between the two datasets.
The camera equipment of the BBC is clearly of higher quality than the webcam of the Hessian Parliament so that despite nominally the same video resolution, there is a difference in quality between the video datasets due to dissimilar dynamic range of the camera sensor, possibly existing camera-internal post-processing, as well as more elaborately calculated and manufactured lenses. In addition, external factors such as shorter camera distance (object distance), the partially existing professional and intelligibility-oriented speech training of the news presenters and the more professional lighting in the BBC dataset provide a clearer, higher-contrast and sharper image of the lip movements, so that it can be expected that LRW-trained models for lip reading will have a higher performance in terms of word recognition than will be the case with GLips. The quality of the audio recordings, which were integrated into the .mp4 format in LRW and are available separately as .m4a in GLips, is less deviant due to the use of high-quality microphones in the Hessian Parliament. However, for training our lip reading models we will only use the visual information. Furthermore, the number of speakers in LRW is several hundred, which is significantly higher than in GLips, which is estimated to be around 100.

\section{Model Evaluation}
\label{chapter:evaluation}

\begin{figure}[h]
\centering
\includegraphics[width=0.93\columnwidth]{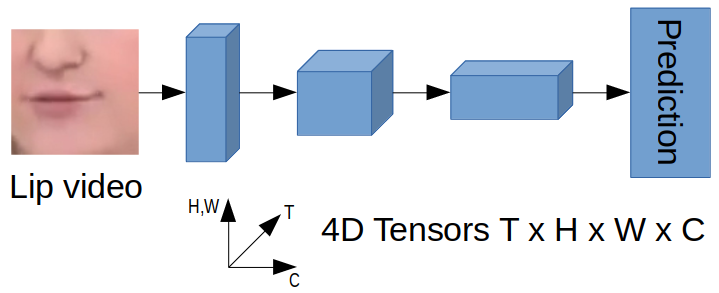} 
\caption{X3D Architecture}
\label{fig.9}

\end{figure}

There are several recurrent (e.g.~\cite{stafylakis2017interspeech}) and feedforward (e.g.~\cite{Weng-2019-118899}, \cite{Martnez2020LipreadingUT}) models for lip reading on word level.
We chose the X3D convolutional neural network model by ~\newcite{feichtenhofer2020x3d}
since it is efficient for video classification in terms of accuracy and computational cost and well designed for the processing of spatiotemporal features.
The model is depicted  in Figure~\ref{fig.9},
has 4D tensors as its main layers, with one dimension each for the temporal dimension (T), height and width (H $\times$ W) of spatial dimensions and number of channels (C).
It is an expanding image processing architecture that uses channel-wise convolutions as building blocks.
Synchronized stochastic gradient descent (SGD) was performed of parallel workers following the linear scaling rule for learning rate and minibatch size to reduce
training time \cite{goyal2018accurate}.

We used the official model implementation\footnote{GitHub - X3D implementation: \url{https://github.com/facebookresearch/SlowFast/blob/main/MODEL\_ZOO.md}} that is included in Pytorch Lightning-Flash\footnote{GitHub - PyTorch Lightning Flash: \url{https://github.com/PyTorchLightning/lightning-flash}}
and for video processing we use the PyTorchVideo~\cite{fan2021pytorchvideo} library.
The model was implemented and tested on a single NVIDIA Geforce RTX2080Ti.

\subsection{Experiments with GLips and LRW}

To evaluate whether the word recognition rate of the lip reading models can be improved by transfer learning, we conducted two experiments. To keep the transfer learning computations manageable, we create two subsets of each of the LRW and GLips datasets, which we call LRW\textsubscript{15} and GLips\textsubscript{15}, and which consist of only 15 randomly selected words of 500 instances each instead of all 500 words. Two further subsets named GLips\textsubscript{15-small} and LRW\textsubscript{15-small} consist only of a total of 95 word instances of the same 15 words as the former subsets. We cropped the videos to 96$\times$96 pixels around the lip region to increase the performance in computation and to improve the focus of learning on the lips.\\

In Experiment 1, we use the largest versions of the GLips\textsubscript{15} and LRW\textsubscript{15} datasets among themselves for transfer learning. We test here whether lip reading abilities can be passed among models trained on the same size datasets in a language-independent manner by comparing the word recognition rate of the transfer-learned models with those learned from scratch.\\

\begin{figure*}[tbh]
\begin{center}

\includegraphics[width=\textwidth]{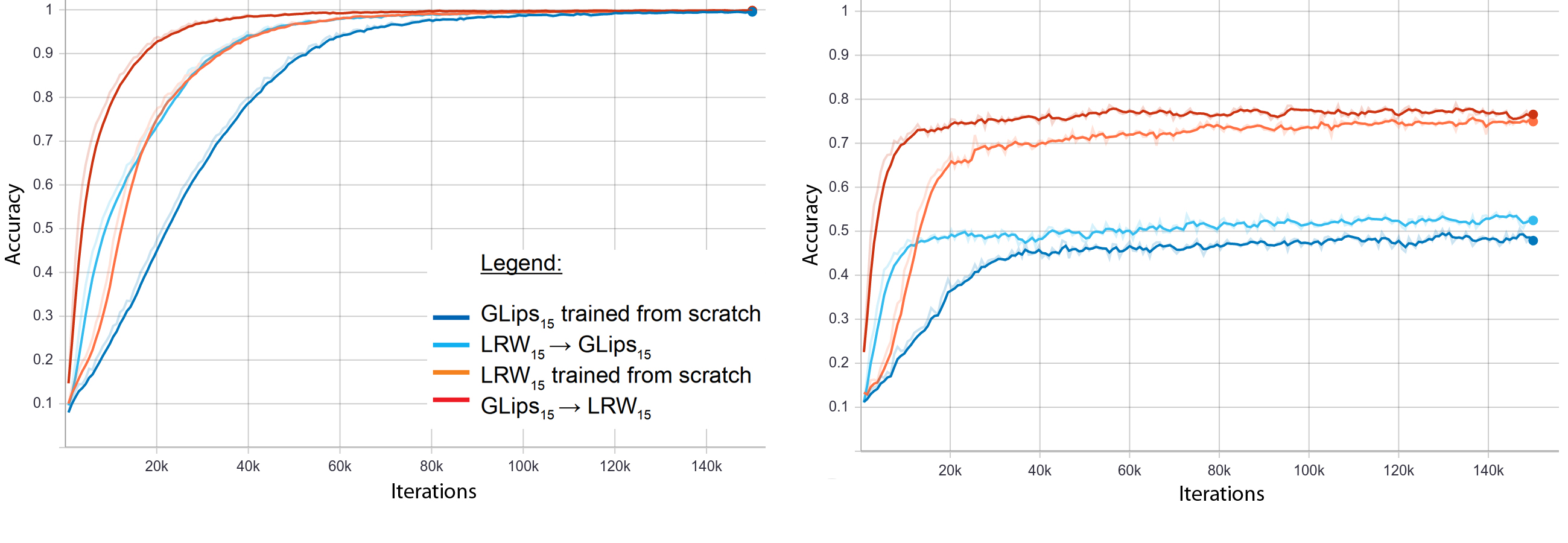} 
\caption{Results of training (left) and validation (right) accuracy per iteration for Experiment 1: transfer learning between same-size datasets}
\label{fig.7}
\end{center}
\end{figure*}

\begin{figure*}[tbh]
\begin{center}

\includegraphics[width=\textwidth]{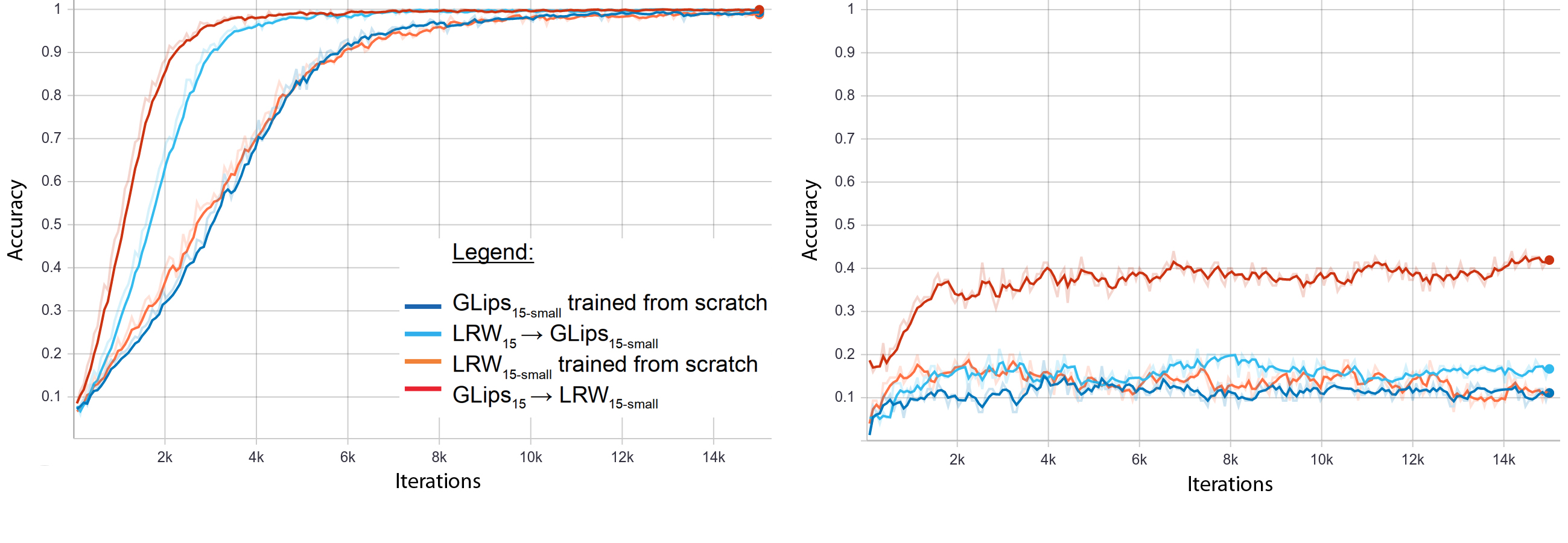} 
\caption{Results of training (left) and validation (right) accuracy per iteration for Experiment 2: transfer learning from a large dataset to a small dataset}
\label{fig.8}
\end{center}
\end{figure*}

In Experiment 2, we test whether the transfer learning benefits are greatest, as theoretically expected, where models learned on large datasets transfer to models learned on smaller datasets by learning from LRW\textsubscript{15}→ GLips\textsubscript{15-small} and also from GLips\textsubscript{15}→ LRW\textsubscript{15-small}, and also by comparing the results to models learned from scratch.

\subsection{Experimental Results}

The smoothened curves of the results were plotted in TensorBoard\footnote{TensorBoard: \url{https://www.tensorflow.org/tensorboard}}.
As seen in Fig.~\ref{fig.7}, in Experiment~1 the validation accuracies of GLips\textsubscript{15} and LRW\textsubscript{15} trained from scratch are lower than those of the transfer-learned models GLips\textsubscript{15}→LRW\textsubscript{15} and LRW\textsubscript{15}→GLips\textsubscript{15}. Also both transfer-learned curves rise steeper in the beginning, which means that the networks learn faster and maintain their higher level across all epochs, manifesting their learning advantage. Additionally, GLips\textsubscript{15}→LRW\textsubscript{15} has a higher starting point, which accelerates the learning rate even more.\\

The results of Experiment 2 in Fig.~\ref{fig.8} also clearly demonstrate
the advantages of transfer learning, but look different in detail.
LRW\textsubscript{15}→GLips\textsubscript{15-small} as well as GLips\textsubscript{15}→LRW\textsubscript{15-small} achieve an advantage over the respective networks trained from scratch. In both experiments the average validation accuracies of the LRW-networks reach a higher score than the GLips-networks. This is particularly pronounced for GLips\textsubscript{15}→LRW\textsubscript{15-small}, which appears has the largest advantage in the validation experiment. It is surprising that GLips as source of transfer learning helps learning LRW more (dark red curve) than vice versa, since GLips as the source of transfer learning has lower-quality videos. An explanation could be that given the low number of data points in this experiment, GLips with its noisier visual features did not allow the model to overfit, while on the other hand, the LRW-pretrained model  (light blue) might have overfit to distinct features in LRW.

\section{Discussion}

The successful transfer learning between the two languages indicates that there are features in both datasets regarding lip reading that are language-independent and thus can be transferred to another language.

Due to the better performance of the LRW-trained networks, we hypothesize that the difference between the models trained on GLips and LRW lies in the quality of the data. In GLips, in comparison, overall learning is subjected to more noise in addition to the features important for the complex task of lip reading. However, the evaluation of the curves shows clear advantages of transfer learning compared to learning from scratch, both in the overall performance and in the speed of learning.
The quality of a dataset is of utmost importance for the performance of a model using it. Transfer learning in neural networks amplifies the effects of the quality differences between datasets. 
In both experiments, we can observe the advantage of the higher quality dataset of LRW for transfer learning, as anticipated in Section 4. In particular, for models trained from small datasets, the benefit of transfer learning compared to learning from scratch becomes substantial.

The benefit of transfer learning for lip reading has recently been shown between English, Romanian and Chinese language \cite{jitaru2021transferlips}.
Their work showcases transfer learning from multiple source languages. Here one needs to take into account that some languages have larger and higher-quality datasets, benefiting transfer learning more than other languages.

Speaker independence according to~\newcite{bear2017visual} is only possible in GLips 
if the dataset would be split into training, validation and test speakers according to different speakers. We did not implement a speaker-independent split in GLips in favor of the larger word set. If splitting GLips in a speaker-independent fashion, the number of words in the training set would reduce to less than 500 for some words.

As a technical implementation, it would be possible to perform a face recognition on the original videos and to link the different identities with the spoken word instances. With this extended information dimension, disjoint test sets can be created with respect to the training and validation sets. Data augmentation like translation or altering the RGB-values of the videos were not used in the training for performance reasons. However, such techniques may be useful on more powerful architectures to reduce the risk of overfitting~\cite{krizhevsky2017imagenet} and to increase generalization performance.

\section{Conclusion}

With GLips we have created the first large-scale German lip reading dataset, which is compatible to the large English LRW dataset. Using an X3D deep neural network, we demonstrated successful transfer learning between the two languages in both directions. This new dataset can support various applications in areas of disability support, communication in noisy environments, boosting of existing ASR systems, etc., progressing state-of-the-art assistive technologies.
Revisiting the publicly available source of the dataset, further applications would become possible, such as learning automatic speech recognition, and extended TextGrid information will allow to create a dataset for sentence-level recognition from the original videos.

\section{Acknowledgements}
We gratefully acknowledge partial support from the German Research Foundation (DFG) under the project Crossmodal Learning (CML, Grant TRR 169).

\balance
\section{Bibliographical References}
\label{reference}
\bibliographystyle{lrec2022-bib}
\bibliography{GLips}

\section{Language Resource References}
\label{lr:ref}
\bibliographystylelanguageresource{lrec2022-bib}
\bibliographylanguageresource{languageresource}

\end{document}